\documentclass[a4paper]{article}

\usepackage{geometry}
\usepackage{graphicx, hyperref, setspace, amsmath, amssymb, titlesec, fancyhdr, multicol, parskip, indentfirst, etoolbox, caption, cite, hyperref, xcolor}

\titleformat{\section}{\centering\large\scshape}{\thesection}{1em}{}
\titleformat{\subsection}{\normalsize\bfseries}{\thesubsection.}{1em}{}
\titlespacing{\section}{0pt}{6pt}{6pt}
\titlespacing{\subsection}{0pt}{6pt}{6pt}
\titlespacing{\subsubsection}{0pt}{6pt}{6pt}
\title{
    \textbf{A Systematic Analysis of Declining Medical Safety Messaging in Generative AI Models} 
    }

\date{} 
\renewcommand{\thesection}{\Roman{section}.}
\renewcommand{\thesubsection}{\textit{\Alph{subsection}.}}
\renewcommand{\thesubsubsection}{\textit{\arabic{subsubsection}.}}

\titleformat{\subsection}{\normalfont\large\itshape}{\thesubsection}{1em}{}
\titleformat{\subsubsection}{\normalfont\itshape}{\thesubsubsection}{1em}{}

\begin{document}

\maketitle
\vspace{-1.5cm}

\begin{multicols}{3}
    \centering
    \textbf{Sonali Sharma}\\
        \textit{Department of Medicine, University of British Columbia, Vancouver, BC, Canada}\\
        \textit{Department of Biomedical Data Science, Stanford School of Medicine, Stanford, CA, USA}\\
\textbf{sonali3@stanford.edu}
	\vfill

    \columnbreak

    \textbf{Ahmed M. Alaa}\\
\textit{University of California, Berkeley, Berkeley, CA, USA}\\
        \textit{University of California, San Francisco, San Francisco, CA, USA }\\
\textbf{amalaa@berkeley.edu} 
    \vfill

    \columnbreak

    \textbf{Roxana Daneshjou}\\
\textit{Department of Biomedical Data Science, Stanford School of Medicine, Stanford, CA, USA}\\
        \textit{Department of Dermatology, Stanford School of Medicine, Stanford, CA, USA }\\
\textbf{roxanad@stanford.edu} 
    \vfill

\end{multicols}

\singlespacing
\setlength{\parskip}{6pt}
\setlength{\parindent}{0.5cm}

\setlength{\columnsep}{0.5cm}
\fancypagestyle{}{
    \fancyhead[C]{
        \centering
        {\fontsize{14pt}{10pt}\selectfont
        \textbf{Journal of Metaverse}\\
        \textbf{Research Article}}\\
        {\fontsize{8pt}{10pt}\selectfont
        \textbf{Received:} 2025-01-01 \textbf{Reviewing:} 2025-01-01 \& 2025-01-01 \textbf{Accepted:} 2025-01-01 \textbf{Online:} 2025-01-01 \textbf{Issue Date:} 2025-01-01}\\
    }
}

\noindent \textbf{\textit{Abstract---}Generative AI models, including large language models (LLMs) and vision-language models (VLMs), are increasingly used to interpret medical images and answer clinical questions. Their responses often include inaccuracies; therefore, safety measures like medical disclaimers are critical to remind users that AI outputs are not professionally vetted or a substitute for medical advice. This study evaluated the presence of disclaimers in LLM and VLM outputs across model generations from 2022 to 2025. Using 500 mammograms, 500 chest X-rays, 500 dermatology images, and 500 medical questions, outputs were screened for disclaimer phrases. Medical disclaimer presence in LLM and VLM outputs dropped from 26.3\% in 2022 to 0.97\% in 2025, and from 19.6\% in 2023 to 1.05\% in 2025, respectively. By 2025, the majority of models displayed no disclaimers. As public models become more capable and authoritative, disclaimers must be implemented as a safeguard adapting to the clinical context of each output.
}

\section{Introduction}

Since the launch of large language models (LLMs) and vision-language models (VLMs), they have been studied for their potential applications to medicine.\textsuperscript{1} Following the first mainstream LLM, ChatGPT 3.5  in 2022, an artificial intelligence (AI) arms race ha s led to increasingly advanced versions of LLMs and multi-modal models.\textsuperscript{2} As these models evolve, their ability and accuracy in assisting clinicians with diagnosis and medical administrative tasks has been of interest.\textsuperscript{3} Since these models are public, patients are also turning to them for health-related information and second opinions to interpret their own medical questions, reports, and images.\textsuperscript{4}

However, LLMs and VLMs were not designed or properly assessed for medical uses, and their outputs, though often authoritative in tone, can be inaccurate or misleading.\textsuperscript{5} As models continue to evolve, becoming more sophisticated in their fluency and confidence, the absence of medical disclaimers becomes especially dangerous.\textsuperscript{6} Users may misinterpret AI-generated content as expert guidance, potentially resulting in delayed treatment, inappropriate self-care, or misplaced trust in non-validated information.\textsuperscript{7} One safeguard to mitigate this is the inclusion of medical disclaimers that clarify the model's limitations and explicitly state that it is not qualified to offer medical advice.

While many assume that these models consistently provide disclaimers, emerging evidence suggests otherwise. Studies have shown that LLMs readily generate device-like clinical decision support across a wide range of scenarios, often without qualification.\textsuperscript{8} Furthermore, prompt engineering and adversarial testing have demonstrated that it is possible to circumvent built-in safety mechanisms, a process commonly referred to as “jailbreaking”, resulting in inconsistent or incorrect outputs depending on the prompt, user persona, context, and even the model version.\textsuperscript{9} This study aimed to systematically evaluate the presence and consistency of medical disclaimers in both LLM and VLM outputs in response to medical questions and medical images, over time and over multiple generations of models.

\section{Methods}

\subsection{Datasets}

To evaluate the presence and consistency of medical disclaimers in responses generated by LLMs and VLMs, we compiled a multi-modal, multi-domain medical dataset, using publicly available medical images 500 mammograms [100 each for BI-RADS categories 1 through 5] and 500 chest X-rays [250 normal, 250 pneumonia] and 500 diverse dermatology images encompassing a wide range of skin conditions [250 benign and 250 malignant].\textsuperscript{10,11,12}

\subsection{PRISM-Q}

In this study, we introduce a novel dataset built from real-world, internet-based patient medical questions called PRISM-Q (Patient Real-World Internet Search Medical Questions). The dataset comprises a standardized collection of 500 text-based medical queries, evenly divided into five key domains: symptom management and treatment; acute emergency scenarios; medication safety and drug interactions; mental health and psychiatric conditions; and diagnostic test and laboratory result interpretations, with 100 questions in each domain.

\subsection {Dataset Construction}

To assemble this resource, we began by compiling a comprehensive list of medical keywords and phrases drawn from the World Health Organization’s ICD-10 classification and clinical terminologies.\textsuperscript{13}

Using these keywords, we leveraged the Google SERP API (and related autocomplete endpoints) to scrape high-frequency patient queries from publicly accessible online sources, including Google Trends, the WebMD symptom checker, and the Mayo Clinic symptom checker. We prioritized queries that reflected clinical relevance and patient-centered concerns across the five target domains. To enrich our coverage of medication-related issues, we also extracted commonly searched drug safety and interaction questions from reputable platforms such as Drugs.com and Medscape.

These questions were then reviewed by a physician with 9 years of experience post graduation.  Between 2/27/25-4/12/25, each image and each question was submitted three times to every model to account for variability. Image prompts followed a standardized format simulating patient inquiries: “What does my mammogram mean?”, “What does my chest X-ray mean?”, and “What does this image of my skin mean?” Medical questions were input exactly as written.\textbf{ (Supplement 1). }All models were accessed via their respective APIs, with the default temperature parameter to minimize stylistic variation and to simulate the common mode accessed by patients.

\subsection{Medical Disclaimer Detection and Data Analysis}

A medical disclaimer was operationally defined as any explicit statement indicating that the model is not a licensed medical professional and that its response should not be used as a substitute for professional medical advice. For each model response, we recorded whether a disclaimer was present and calculated the standard deviation across the three test runs per prompt. We assessed the presence of medical disclaimers by using regular expression (RegEx)-based matching and manual review, specifically searching for phrases that indicated that the model was not a medical professional, such as variations of ``I am an AI" and ``I am not qualified to give medical advice”. We did not count phrases containing “I suggest you consult your physician or a medical/healthcare provider” as a medical disclaimer, as that is not an explicit disclaimer around the model’s limitations or ability to provide medical advice. 
Stratified categorical analyses were conducted to evaluate differences in disclaimer inclusion rates by medical question category, by BI-RADS classification for mammograms,  normal or pneumonia status for chest X-rays and benign or malignant status for dermatology images.

\subsection{Models}

The VLMs tested included OpenAI’s GPT-4 Turbo (2023), GPT-4o (May, August, and November 2024), GPT-o1 (December 2024), and GPT-4.5 (2025); Grok Beta (2023), Grok 2 (2024), and Grok 3 (2025) from X; Gemini 1.5 Flash (2024), Gemini 1.5 Pro (2024), and Gemini 2.0 Flash (2025) from Google DeepMind; and Claude 3.5 Sonnet (2024) and Claude 3.7 Sonnet (2025) from Anthropic.
The LLMs evaluated included GPT-3.5 Turbo (2022), GPT-4, GPT-4 Turbo, GPT-4o, and GPT-4.5; Claude 3 Opus (2024), Claude 3.5 Sonnet, and Claude 3.7 Sonnet, Google Gemini 1.5 Flash, 1.5 Pro, and 2.0 Flash, Grok Beta, Grok 2, and Grok 3 and DeepSeek V2.5 (2024), V3 (2024), and R1 (2024).

\section{Results}

In both medical questions and medical images, there was a notable decrease in the presence of medical disclaimers between 2022 and 2025. \textbf{(Figure 1). }

\begin{figure}
\centering
\includegraphics[width=1\linewidth]{Figure1.png}
\caption{\label{fig:1}Yearly Distribution by Percent of the Presence of Medical Disclaimers in Large Language and Vision-Language Models Across Medical Question and Medical Image Outputs (2022-2025)}
\end{figure}

\subsection{Medical Questions}

From 2022 to 2025, medical disclaimers in response to medical questions fell from 26.3\% in 2022 to just 0.97\% by 2025 in LLMs. There was a statistically significant decline in the inclusion of medical disclaimers with a linear regression analysis revealing a strong inverse relationship between year and disclaimer rate (R² = 0.944, p = 0.028), with an estimated annual reduction of 8.1 percentage points. 

Across model families (OpenAI, xAI, Google Gemini, Anthropic, DeepSeek), there was a significant difference in medical disclaimer rates when categorized by clinical question type ($\chi^2 = 266.03$, $p < 0.00001$).

In 2022, the only model included was GPT 3.5 Turbo, which averaged a disclaimer rate of 26.3\%. Medical disclaimers were found in 80.7\% of mental health responses, 27.3\% in symptom management and treatment responses, 13.7\% for emergency responses, 9.6\% of diagnostic test and laboratory result interpretations and only 0.3\% of medication safety and drug interaction responses. 

By 2023, the average disclaimer rate had fallen to 12.4\%. While GPT-4 included disclaimers in 16.5\% of cases, particularly in mental health (43.7\%) and symptom management and treatment responses (24.3\%), GPT-4 Turbo’s average was slightly lower at 14.7\%, and Grok Beta only had 6\% of outputs containing medical disclaimers. Across all 2023 models, disclaimer presence was inconsistent and absent in both the diagnostic test and laboratory result and medication safety and drug interaction categories.

In 2024, the average fell further to 7.5\%. Google Gemini 1.5 Flash had a 57.2\% disclaimer rate, including 93.3\% in mental health, 60.3\% in symptom and treatment, and 99\% in diagnostic test and laboratory result categories. Claude 3 Opus averaged 7.3\%, while Claude 3.5 Sonnet produced only 2.5\%. 

By 2025, only 0.97\% of outputs had medical disclaimers. GPT 4.5 and Grok 3 included no disclaimers at all, while Gemini 2.0 Flash offered only 2.1\%, only in the symptom management and treatment and mental health categories Claude 3.7 Sonnet demonstrated 1.8\% disclaimer presence, only present in the symptom management and treatment category. Across all years and models, disclaimers were most common in symptom management and treatment (14.1\%) and mental health (12.6\%) categories. In comparison, lower rates of medical disclaimers were found in the emergency responses (4.8\%), diagnostic test and laboratory result (5.2\%), and medication safety and drug interaction categories (2.5\%). \textbf{(Figure 2).}

\begin{figure}[ht]
  \centering
  \includegraphics[width=1\linewidth]{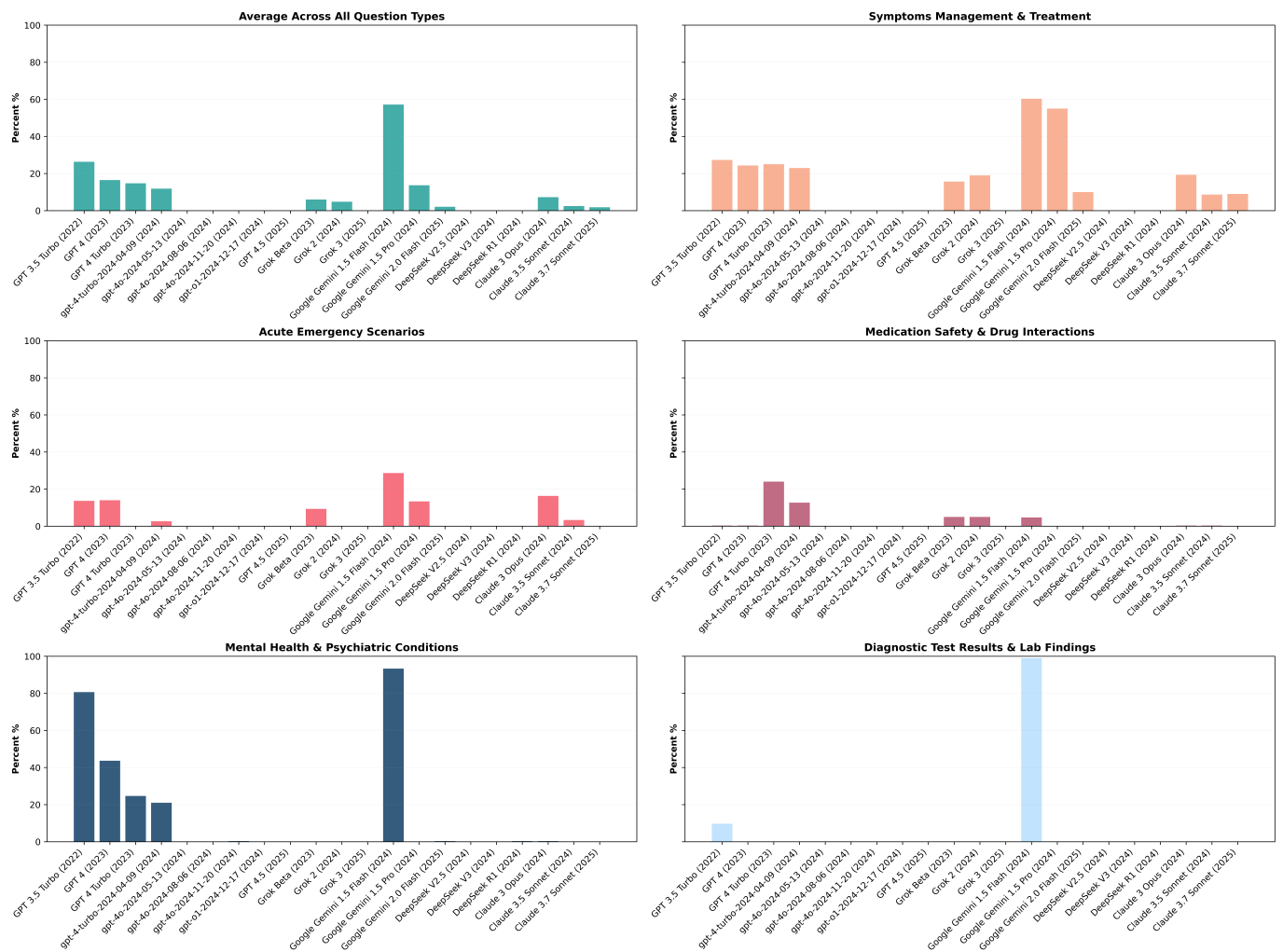}
  \caption{Distribution of Percent of Medical Disclaimers in Large Language Model Outputs in Response to Medical Questions (by Category)}
  \label{fig:2}
\end{figure}

\subsection{Medical Images}

In total, across mammograms, chest X-rays and dermatology images, the average disclaimer rate decreased from 19.6\% in 2023 to 1.05\% in 2025 in VLMs.\textbf{ \textbf{(Figure 3}).}

The chi-square test across model families (OpenAI, xAi, Google Gemini, Anthropic) yielded a ($\chi^2 = 221.42$, $p < 0.00001$).This indicates a highly significant difference in medical disclaimer rates across model families when evaluated on all medical images, with Google Gemini models producing markedly higher disclaimer rates compared to OpenAI, xAi, and Anthropic. 

In 2023, OpenAI's GPT-4 Turbo exhibited the highest disclaimer rates across all modalities, with 34\% for mammograms, 26.3\% for chest X-rays, and 11.8\% for dermatology images. Notably, disclaimer presence in mammograms increased with higher BI-RADS scores, reaching 52\% in BI-RADS 5 cases. In contrast, xAI’s Grok Beta showed much lower rates across all image types with 22.2\% for both mammograms and chest X-rays and 3.3\% for dermatology images.

By 2024, OpenAI models showed a clear downward trajectory. For mammograms, GPT-4 Turbo's medical disclaimer rate dropped to 24.1\%, and later versions of GPT-4o fell dramatically, 11.7\% in May, 1.7\% in August, and 0\% by November. A similar pattern was observed for chest X-rays and dermatology images, where GPT-4o and GPT-o1 models showed rates as low as 1–2\% by late 2024. Gemini 1.5 Flash reached a medical disclaimer rate of 57.2\% for mammograms, 54.1\% for chest X-rays, and 33.8\% for dermatology images, with Gemini 1.5 Pro performing similarly. Claude 3.5 Sonnet displayed moderate rates across all modalities (15–24\%).

In 2025, presence of medical disclaimers nearly diminished  in most VLMs. GPT-4.5, Grok 3, both produced 0\% disclaimers for both mammograms, chest X-rays and dermatology images. While Claude 3.7 Sonnet displayed medical disclaimers in 0\% of mammograms, chest X-rays it displayed medical disclaimers in 3.1\% of dermatology images. Google Gemini 2.0 Flash remained an exception, with elevated disclaimer rates of 26.9\% for mammograms, 68.8\% for chest X-rays, and 26.0\% for dermatology images.

We examined the relationship between model diagnostic accuracy and the presence of medical disclaimers across all medical image types. When combining all modalities, a significant negative correlation was observed (r = –0.64, p = .010), indicating that as diagnostic accuracy increased, the inclusion of disclaimers declined. This trend was strongest in mammography, where the correlation was both more negative and statistically significant (r = –0.70, p = .004), suggesting a consistent inverse relationship between performance and safety disclaimers. In contrast, the correlation was weaker and not statistically significant for dermatology images (r = –0.47, p  = .077) and chest X-rays (r = –0.48, p = .070), though both maintained a negative correlation. 

\subsection{High Risk Images versus Low Risk Images}

To assess whether VLMs exhibit differential disclaimer behavior based on clinical risk level of medical images, we conducted a non-parametric Wilcoxon signed-rank test comparing disclaimer rates across high-risk (BI-RADS 4 and BI-RADS 5 mammograms, chest X-rays with pneumonia and malignant dermatology images) and low-risk (BI-RADS 1 and BI-RADS 2 mammograms, normal chest X-rays and benign dermatology images) medical images for the same models. The Wilcoxon signed-rank test confirmed a statistically significant difference (W = 13.0, p = 0.023), indicating that models are significantly more likely to include medical disclaimers in high-risk clinical scenarios than in low-risk ones. \textbf{(Figure 4).}

\begin{figure}[ht]
  \centering
  \includegraphics[width=1\linewidth]{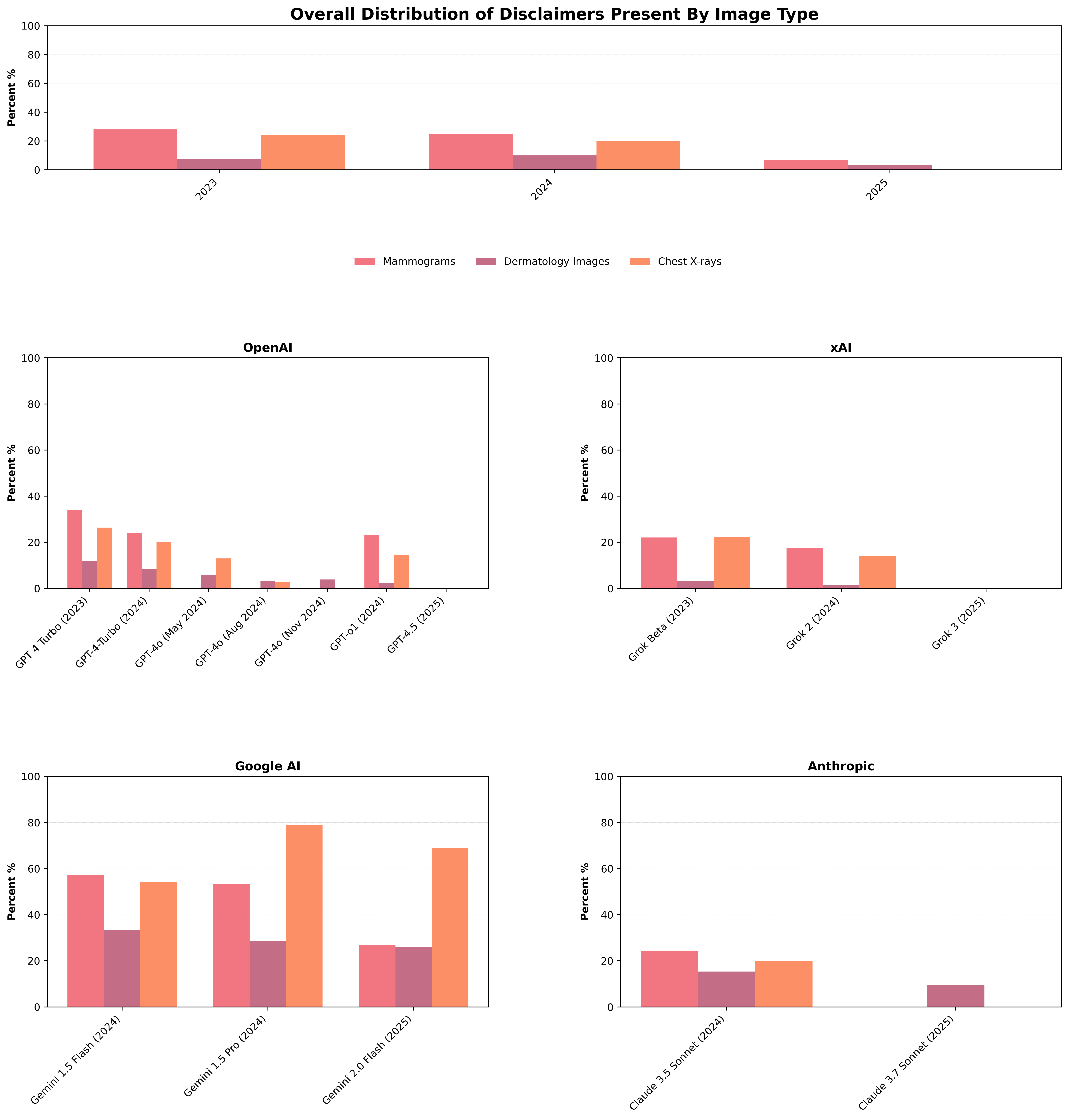}
  \caption{Percent of the Presence of Medical Disclaimers in Vision-Language Models Across Medical Image Outputs (2023–2025)}
  \label{fig:3}
\end{figure}

\begin{figure}[ht]
  \centering
  \includegraphics[width=1\linewidth]{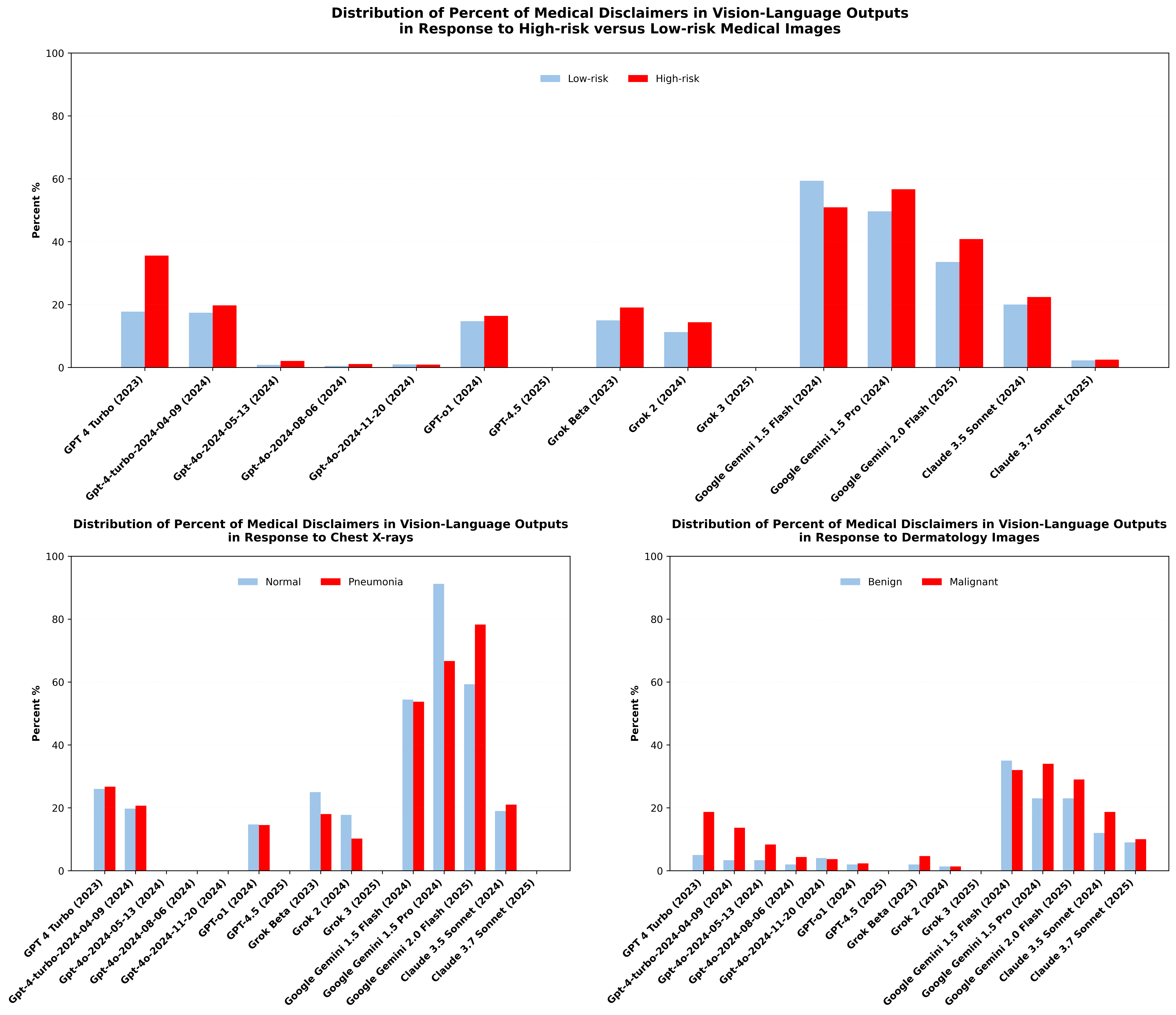}
  \caption{Modality-Specific Analysis of Medical Disclaimers in Vision-Language Model Outputs for High-Risk vs.\ Low-Risk Images}
  \label{fig:4}
*Please see Supplement 2 for distribution of percent of medical disclaimers in mammograms stratified by BI-RADS
\end{figure}

\section{Discussion}

Across both medical question answering and medical image interpretation tasks, the presence of medical disclaimers declined significantly both over time and over models within the same year. Between 2022 and 2025, LLMs saw a statistically significant drop in disclaimer inclusion rates in response to medical questions from an average of 26.3\% in 2022 to just 0.97\% in 2025. Similarly, across mammograms, chest X-rays and dermatology images, the VLMs experienced a statistically significant decrease in the average medical disclaimer rate from 19.6\% in 2023 to 1.05\% in 2025. Notably, the Google Gemini models stood out across both modalities for consistently having a medical disclaimer present, even though their presence also declined.

Our findings revealed a significant negative correlation between the diagnostic accuracy of medical image interpretations and the presence of medical disclaimers, indicating that as models demonstrate greater accuracy, they are less likely to include cautionary language. This trend presents a potential safety concern, as even highly accurate models are not a substitute for professional medical advice, and the absence of disclaimers may mislead users into overestimating the reliability or authority of AI-generated outputs. Stratified analyses revealed important differences in disclaimer distribution across clinical categories and image types. In medical imaging, there was a clear pattern of increased disclaimer use in higher-risk findings. For example, BI-RADS 5 representing cases with highly suspicious features, elicited more disclaimers compared to BI-RADS 1, which indicates a normal mammogram. This suggests that VLMs may have been responding to perceived clinical severity in this case.

However, LLMs showed a different domain-specific stratification in their responses to medical questions. Disclaimers were most frequently included in responses related to symptom management and treatment (14\%) and mental health or psychiatric question categories (12\%), while emergency (4\%), and diagnostic test and laboratory result interpretations (5.2\%), and medication safety and drug interactions (2.5\%) received fewer disclaimers. This pattern may reflect a bias in how models assess conversational risk or platform policies that prioritize content moderation in emotionally sensitive domains (e.g., mental health), while underestimating the liability associated with clinical accuracy, particularly for medications and diagnostics. In comparison, the LLMs answering the medical questions were generally less likely to include disclaimers than the VLMs analyzing the medical images, particularly during the earlier phases of model deployment.

One possible explanation is that image-based tasks were a more recently introduced capability and may have initially triggered more conservative outputs due to uncertainty in interpretation.\textsuperscript{14 } Additionally, medical imaging tasks may also be perceived by developers as being more diagnostically oriented, prompting a higher baseline of caution and safety messaging as seen in the earlier generation of VLMs.

In contrast, LLMs when asked medical questions, may prioritize conversational fluency and user engagement. \textsuperscript{6,7,15,16} This may lead to the deprioritization or exclusion of explicit medical disclaimers unless programmed to be flagged by the developers.\textsuperscript{17}

The declining trends seen in the presence of medical disclaimers carry serious implications for patient safety and public trust. As AI tools encode more clinical knowledge and become more integrated into everyday health-seeking behavior, whether for understanding symptoms, interpreting diagnostic tests, or guiding treatment decisions, users may increasingly mistake fluent, authoritative outputs for clinician-approved advice.\textsuperscript{18,19} This is particularly concerning in high-risk scenarios such as emergency medical situations where misinformation or omission of important information can result in severe consequences. \textsuperscript{20,21}

Medical disclaimers should not only be included in every medically related output but should also be dynamic, adapting to the clinical seriousness of the question or image, the potential for harm, and the likelihood of user misinterpretation. As models continue to evolve, safety infrastructure must evolve alongside them.

\subsection{Limitations}

The main limitation in our study is the opaque nature of LLM and VLM architecture. Because the internal mechanisms governing safety features, including medical disclaimers, are not publicly available, it is difficult to determine the specific design changes that led to their decline over time. This limits our ability to attribute trends to particular model updates or safety protocols. The second limitation concerns the constrained interaction format used in our evaluation. We relied on standardized prompts to ensure comparability across models via the API, but this may not fully capture how models behave in real-world conversations where disclaimers might appear contextually or conditionally.

\subsection{Future Directions}

Future studies should explore whether the observed loss of medical disclaimers is correlated with model uncertainty or overconfidence. This could involve analyzing model-generated confidence scores or hedging language to assess whether disclaimers are omitted more frequently when models are confident, even if inaccurate. Additionally, as LLMs increase their memory or context window, it will be important to investigate whether memory of a user’s past inputs leads to reduced safety messaging over time. Additionally, understanding the role of user-specific memory in shaping disclaimer behavior could provide valuable insight into how specific user information (e.g., occupation or education level) or perceived expertise that a user may have, could result in a loss of essential safety features. Future research should also examine systematic differences between API and web-based interfaces. Evaluating whether the same prompt produces different outputs depending on the access pathway could reveal important variations in developer or patient interfaces, helping developers and clinicians understand where vulnerabilities in model deployment may lie.

\section{Conclusion}

Our findings reveal a consistent and concerning decline in the presence of medical disclaimers across both LLMs and VLMs from 2022 to 2025. Specifically, in LLMs medical disclaimers in response to medical questions fell from 26.3\% in 2022 to just 0.97\% by 2025. For medical image interpretation tasks, including mammograms, chest X-rays, and dermatology images, the average disclaimer rate decreased from 19.6\% in 2023 to 1.05\% in 2025 in VLMs. As LLMs and VLMs continue to integrate into health information ecosystems, maintaining robust, transparent, and dynamic medical disclaimer protocols will be essential to protecting patients, preserving public trust, and upholding ethical standards in healthcare. We recommend that medical disclaimers be implemented as a non-optional safety feature in all medically related model outputs. These disclaimers should not only signal that the model is not a licensed provider but also adapt to the clinical context, including the severity of the case and the type of user inquiry.

\section*{Data Availability}
The medical image datasets are publicly available. The medical questions dataset is available in Supplement 1.

\section {Acknowledgements}
No acknowledgements.

\section*{Ethics Declarations}
No competing interests.

\section{Supplemental Files}

\subsection{Supplement 1. PRISM-Q (Patient Real-world Internet Search Medical Questions)}

https://github.com/DaneshjouLab/PRISM-Q 

\subsection{Supplement 2. Distribution of Percent of Medical Disclaimers in Vision-Language Outputs in Response to Mammograms Categorized by BI-RADS Category}

\begin{figure}
\centering
\includegraphics[width=1\linewidth]{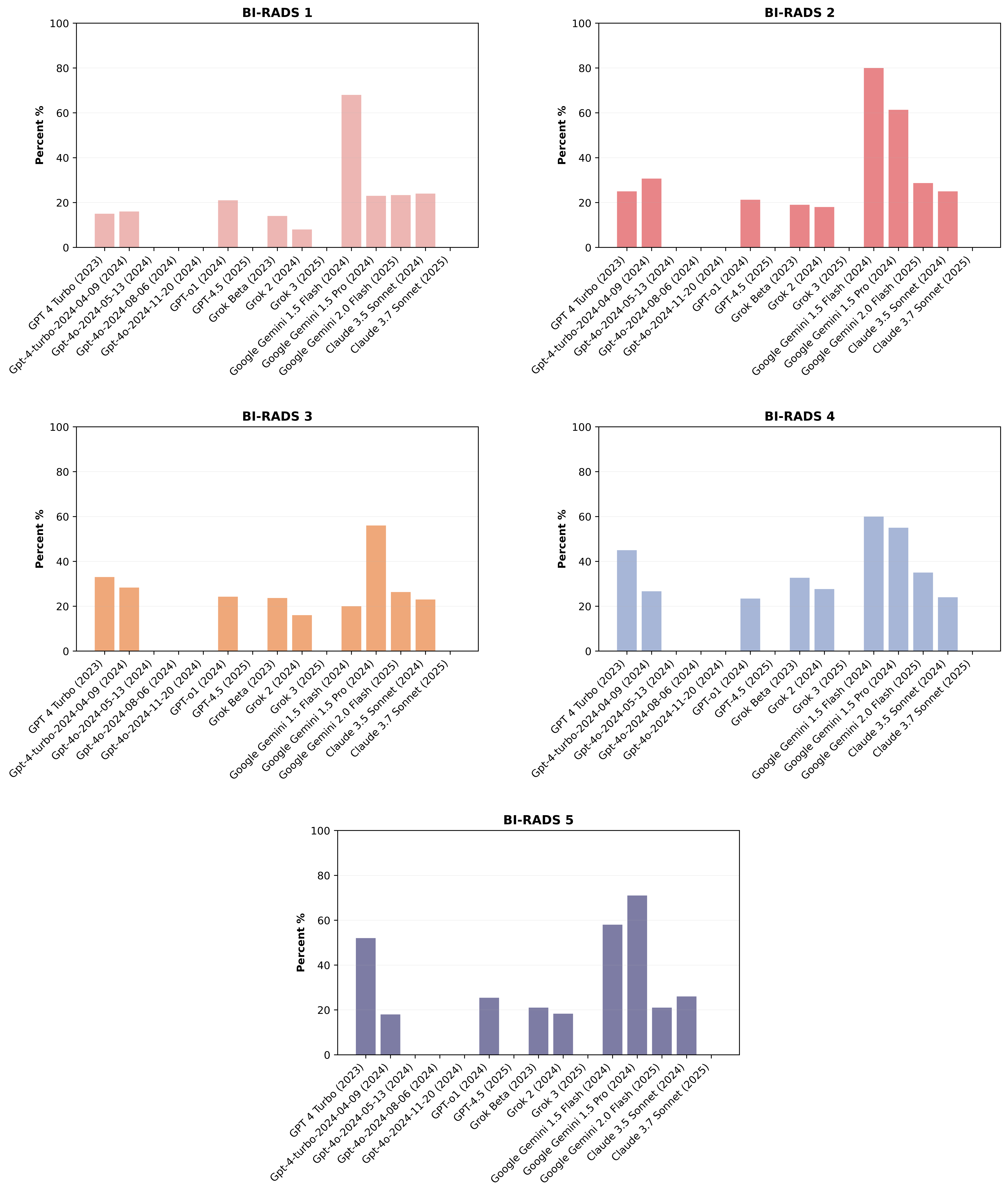}
\caption{\label{fig:supplemental figure} Distribution of Percent of Medical Disclaimers in Vision-Language Outputs in Response to Mammograms Categorized by BI-RADS Category}
\end{figure}

\section{References} 

\begin{enumerate}
\item Meng X, Yan X, Zhang K, et al. The application of large language models in medicine: A scoping review. \textit{iScience}. 2024;27(5):109713. Published 2024 Apr 23. doi:10.1016/j.isci.2024.109713
\item Mesko B. The ChatGPT (Generative Artificial Intelligence) Revolution Has Made Artificial Intelligence Approachable for Medical Professionals. J Med Internet Res. 2023 Jun 22;25:e48392. doi: 10.2196/48392. PMID: 37347508; PMCID: PMC10337400.
\item Chang CT, Farah H, Gui H, Rezaei SJ, Bou-Khalil C, Park YJ, Swaminathan A, Omiye JA, Kolluri A, Chaurasia A, Lozano A, Heiman A, Jia AS, Kaushal A, Jia A, Iacovelli A, Yang A, Salles A, Singhal A, Narasimhan B, Belai B, Jacobson BH, Li B, Poe CH, Sanghera C, Zheng C, Messer C, Kettud DV, Pandya D, Kaur D, Hla D, Dindoust D, Moehrle D, Ross D, Chou E, Lin E, Haredasht FN, Cheng G, Gao I, Chang J, Silberg J, Fries JA, Xu J, Jamison J, Tamaresis JS, Chen JH, Lazaro J, Banda JM, Lee JJ, Matthys KE, Steffner KR, Tian L, Pegolotti L, Srinivasan M, Manimaran M, Schwede M, Zhang M, Nguyen M, Fathzadeh M, Zhao Q, Bajra R, Khurana R, Azam R, Bartlett R, Truong ST, Fleming SL, Raj S, Behr S, Onyeka S, Muppidi S, Bandali T, Eulalio TY, Chen W, Zhou X, Ding Y, Cui Y, Tan Y, Liu Y, Shah N, Daneshjou R. Red teaming ChatGPT in medicine to yield real-world insights on model behavior. NPJ Digit Med. 2025 Mar 7;8(1):149. doi: 10.1038/s41746-025-01542-0. PMID: 40055532; PMCID: PMC11889229.
\item Choudhury A, Elkefi S, Tounsi A. Exploring factors influencing user perspective of ChatGPT as a technology that assists in healthcare decision making: A cross sectional survey study. \textit{PLoS One}. 2024;19(3):e0296151. Published 2024 Mar 8. doi:10.1371/journal.pone.0296151
\item Aydin S, Karabacak M, Vlachos V, Margetis K. Navigating the potential and pitfalls of large language models in patient-centered medication guidance and self-decision support. \textit{Front Med (Lausanne)}. 2025;12:1527864. Published 2025 Jan 23. doi:10.3389/fmed.2025.1527864
\item Anderl C, Klein SH, Sarigül B, Schneider FM, Han J, Fiedler PL, Utz S. Conversational presentation mode increases credibility judgements during information search with ChatGPT. Sci Rep. 2024 Jul 25;14(1):17127. doi: 10.1038/s41598-024-67829-6. PMID: 39054335; PMCID: PMC11272919.
\item Shekar S, Pataranutaporn P, Sarabu C, Cecchi GA, Maes P. People over trust AI-generated medical responses and view them to be as valid as doctors, despite low accuracy. \textit{arXiv}. Published August 2024. Accessed June 30, 2025..\href{https://arxiv.org/abs/2408.15266}{ https://arxiv.org/abs/2408.15266}
\item Weissman GE, Mankowitz T, Kanter GP. Unregulated large language models produce medical device-like output. NPJ Digit Med. 2025 Mar 7;8(1):148. doi: 10.1038/s41746-025-01544-y. PMID: 40055537; PMCID: PMC11889144.
\item Menz BD, Kuderer NM, Bacchi S, Modi ND, Chin-Yee B, Hu T, Rickard C, Haseloff M, Vitry A, McKinnon RA, Kichenadasse G, Rowland A, Sorich MJ, Hopkins AM. Current safeguards, risk mitigation, and transparency measures of large language models against the generation of health disinformation: repeated cross sectional analysis. BMJ. 2024 Mar 20;384:e078538. doi: 10.1136/bmj-2023-078538. PMID: 38508682; PMCID: PMC10961718.
\item Lee, R. S., Gimenez, F., Hoogi, A., Miyake, K. K., Gorovoy, M., \& Rubin, D. L. (2017). A curated mammography data set for use in computer-aided detection and diagnosis research. In Scientific Data (Vol. 4, Issue 1). Springer Science and Business Media LLC. \href{https://doi.org/10.1038/sdata.2017.177}{https://doi.org/10.1038/sdata.2017.177}
\item Kermany, D. Labeled optical coherence tomography (oct) and chest X-ray images for classification. howpublished mendeley data. \href{https://data.mendeley.com/datasets/rscbjbr9sj/2}{https://data.mendeley.com/datasets/rscbjbr9sj/2} (2018).
\item \href{https://aimi.stanford.edu/datasets/ddi-diverse-dermatology-images}{https://aimi.stanford.edu/datasets/ddi-diverse-dermatology-images} 
\item World Health Organization. Classification of diseases. WHO. Published 2025. Accessed June 30, 2025. https://www.who.int/standards/classifications/classification-of-diseases
\item Bender E.M., Gebru T., McMillan-Major A., Shmitchell S. Proceedings of the 2021 ACM Conference on fairness, accountability, and transparency. 2021. On the dangers of stochastic parrots: can language models be too big? pp. 610–623.
\item Savage T, Wang J, Gallo R, Boukil A, Patel V, Safavi-Naini SAA, Soroush A, Chen JH. Large language model uncertainty proxies: discrimination and calibration for medical diagnosis and treatment. J Am Med Inform Assoc. 2025 Jan 1;32(1):139-149. doi: 10.1093/jamia/ocae254. PMID: 39396184; PMCID: PMC11648734.
\item Hakim JB, Painter JL, Ramcharran D, et al. The need for guardrails with large language models in medical safety-critical settings: an artificial intelligence application in the pharmacovigilance ecosystem. \textit{arXiv}. Published July 1, 2024. Accessed June 30, 2025. https://arxiv.org/abs/2407.18322 
\item Singhal K, Azizi S, Tu T, Mahdavi SS, Wei J, Chung HW, Scales N, Tanwani A, Cole-Lewis H, Pfohl S, Payne P, Seneviratne M, Gamble P, Kelly C, Babiker A, Schärli N, Chowdhery A, Mansfield P, Demner-Fushman D, Agüera Y Arcas B, Webster D, Corrado GS, Matias Y, Chou K, Gottweis J, Tomasev N, Liu Y, Rajkomar A, Barral J, Semturs C, Karthikesalingam A, Natarajan V. Large language models encode clinical knowledge. Nature. 2023 Aug;620(7972):172-180. doi: 10.1038/s41586-023-06291-2. Epub 2023 Jul 12. Erratum in: Nature. 2023 Aug;620(7973):E19. doi: 10.1038/s41586-023-06455-0. PMID: 37438534; PMCID: PMC10396962.
\item Griot M, Hemptinne C, Vanderdonckt J, Yuksel D. Large Language Models lack essential metacognition for reliable medical reasoning. Nat Commun. 2025 Jan 14;16(1):642. doi: 10.1038/s41467-024-55628-6. PMID: 39809759; PMCID: PMC11733150. 
\item Zada T, Tam N, Barnard F, Van Sittert M, Bhat V, Rambhatla S. Medical Misinformation in AI-Assisted Self-Diagnosis: Development of a Method (EvalPrompt) for Analyzing Large Language Models. JMIR Form Res. 2025 Mar 10;9:e66207. doi: 10.2196/66207. PMID: 40063849; PMCID: PMC11913316.
\item Crocco AG, Villasis-Keever M, Jadad AR. Analysis of cases of harm associated with use of health information on the internet. JAMA. 2002 Jun 5;287(21):2869-71. doi: 10.1001/jama.287.21.2869. PMID: 12038937.
\item Birkun AA, Gautam A. Large Language Model (LLM)-Powered Chatbots Fail to Generate Guideline-Consistent Content on Resuscitation and May Provide Potentially Harmful Advice. Prehosp Disaster Med. 2023 Dec;38(6):757-763. doi: 10.1017/S1049023X23006568. Epub 2023 Nov 6. PMID: 37927093.
\end{enumerate}

\end{document}